\documentclass[11pt,a4paper]{article}

\usepackage[utf8]{inputenc}
\usepackage[T1]{fontenc}

\usepackage[hyperref]{acl2017}
\usepackage{times}
\usepackage{latexsym}

\usepackage{url}

\usepackage{hyperxmp}
\hypersetup{
pdfauthor={Johan Hasselqvist, Niklas Helmertz, Mikael Kågebäck},
pdftitle={Query-Based Abstractive Summarization Using Neural Networks},
pdfsubject={},
pdfkeywords={},
pdfproducer={},
pdfcreator={}
}

\usepackage{amsmath}                
\usepackage{amssymb}                
\usepackage{graphicx} 
\usepackage{tabularx}
\usepackage{booktabs}

\usepackage{xspace}
\usepackage{float} 									

\usepackage{afterpage}

\newcommand{\larrowover}[1]{\stackrel{\leftarrow}{#1}}
\newcommand{\rarrowover}[1]{\stackrel{\rightarrow}{#1}}

\DeclareMathOperator{\GRU}{GRU}

\newcommand{\sub}[1]{_\text{#1}}
\renewcommand{\sup}[1]{^\text{#1}}

\newcommand{\unk}{\texttt{<UNK>}\xspace}

\newcommand{\go}{\texttt{<GO>}\xspace}
\newcommand{\eos}{\texttt{<EOS>}\xspace}

\newcommand{\R}{\mathbb{R}}
\newcommand{\argmax}[1]{\underset{#1}{\operatorname{arg}\,\operatorname{max}}\;}

\newcommand{\Sec}[1]{Section \ref{#1}}

\aclfinalcopy 

\title{Query-Based Abstractive Summarization Using Neural Networks}

\author{Johan Hasselqvist \\
  {\tt \href{mailto:johan@hasselqv.ist}{\color{black}{johan@hasselqv.ist}}} \\\And
  Niklas Helmertz \\
  {\tt \href{mailto:niklas@helmertz.se}{\color{black}{niklas@helmertz.se}}} \\\And
  Mikael Kågebäck\textsuperscript{*}\\
  {\tt \href{mailto:kageback@chalmers.se}{\color{black}{kageback@chalmers.se}}} 
  \AND
  \normalfont
  \textsuperscript{*}%
Department of Computer Science and Engineering,
Chalmers University of Technology
\\
\\
  }

\begin{document}
\maketitle

\begin{abstract}
In this paper, we present a model for generating summaries of text documents with respect to a query. This is known as \emph{query-based summarization}. We adapt an existing dataset of news article summaries for the task and train a pointer-generator model using this dataset. The generated summaries are evaluated by measuring similarity to reference summaries. Our results show that a neural network summarization model, similar to existing neural network models for abstractive summarization, can be constructed to make use of queries to produce targeted summaries.

\end{abstract}
\date{}

\section{Introduction}

Creating short summaries of documents with respect to a query has applications in for example search engines, where it may help inform users of the most relevant results. However, constructing such a summary automatically is a difficult problem yet to be fully solved. 
In this paper, a neural network model for this task is presented. More specifically, the model is designed for brief, commonly single-sentence, summaries. A situation where this may be useful is when a user has performed a search in a search engine and a set of documents have been returned. Concise summaries could then be displayed along with the search results, giving a quick overview of how the document is related to the search query.
What is commonly done in search engines today is that text surrounding an occurrence of a search query in the document is displayed as a summary. This is an example of \emph{extractive} summarization, which produces a summary that only contains parts of the original document. A significant difference in the model we present is that it generates an \emph{abstractive} summary. This type of summary allows for rephrasing and using words not necessarily present in the original document, comparable to a human-written summary. This has the potential of summarizing documents in a more concise way than what is possible with an extractive summary, i.e. making it easier for a reader to understand the relationship between a document and a query.

Automatic text summarization has been a research topic for many years. In general, the goal is to concisely represent the most important information in documents. Much previous work in summarization has been using extractive methods \cite{Nenkova12, mogren2015extractive}. Commonly, individual sentences are extracted and composed together to form a summary. This gives sentences that are as grammatically correct as the source document. They are however inherently limited, and cannot reproduce human-written summaries in general.
Abstractive summarization in particular is closely related to \emph{natural language generation}, and it would be desirable to reach human-level performance in writing summaries. It may however require human-level understanding of the context of documents to produce results comparable to human-written ones.
An important progress in using neural network models for generating text is \emph{sequence-to-sequence}, used by \citet{Sutskever14} for machine translation. It is a way of mapping a varying-length input text to a varying-length output text, and it is applicable to machine translation as well as summarization.
In recent years, progress has been made on using neural network models for text summarization and similar problems. Some examples are sequence-to-sequence models for non-query-based abstractive summarization by \citet{Rush15} and \citet{Nallapati16}. Neural network models have additionally been used for generating image captions \cite{Karpathy15}, which is a form of summary, and for \emph{question answering} problems, such as by \citet{Hermann15} and \citet{Tan15}. Inspired by this progress, we designed a model for query-based summarization using neural networks.

The main contributions of this work includes: (1) A model for query-based abstractive summarization, presented in \Sec{chap:model}. (2) A dataset for query-based abstractive summarization, created by adapting an existing dataset originally used for question answering, described further in \Sec{chap:dataset}. (3) A quantitative evaluation of the performance of the proposed model compared with an extractive baseline and an uninformed abstractive model, presented in \Sec{chap:results}. (4) A qualitative analysis of the generated summaries.

\subsection{Related Work}
\label{chap:related_work}



An early work evaluating several methods for extractive query-based summarization is presented by \citet{Goldstein99}. Besides "full queries", they use "short queries", which on average are 3.9 words. These are similar in length to the types of queries used in the experiments of this thesis work.
Besides the work by \citet{Otterbacher09}, recent work in query-based summarization has been done by \citet{Wang13}, using \emph{parse trees} and \emph{sentence compression}. It is described as not "pure extractive summarization".
During the later stages of this thesis work, \citet{Nema17} propose a neural network model for query-based abstractive summarization, which has some similarities to the model we present. However, the dataset they use is smaller in both average document length and number of documents. Additionally, the types of queries used are different, in that they use complete questions as opposed to our single-entity queries.


The task of \emph{question answering} is to produce an answer to a question posed in natural language. The task is very general and many other problems can be expressed as a question-answering problem. Summarizing with respect to a query may for instance be expressed as "What is a summary of the document with respect to the query X?", for the query X. If the answer to a question is a single complete sentence, then it is especially close to the types of query-based summaries considered in this thesis.
\citet{Otterbacher09} present a model, \emph{Biased LexRank}, which they use for a form of question answering as well as extractive query-based summarization. The answers they generate are full sentences, which makes it similar to our task of query-based summarization.
\citet{Hermann15} present neural network models for question answering. For training these, they create a large dataset from CNN/Daily Mail news articles. We adapt this dataset for query-based summarization, as detailed in Chapter \ref{chap:dataset}. 
\citet{Kumar16} introduce \emph{Dynamic Memory Networks}, which they show reached state-of-the-art performance in a variety of NLP tasks. We draw inspiration from their use of a question module when we incorporate query information in our model.


General abstractive summarization differs from query-based summarization in that a document is summarized without respect to a query.
\citet{Nallapati16} build upon a machine translation model by \citet{Bahdanau15} and generate general abstractive summaries on multiple datasets, including the CNN/Daily Mail dataset by \citet{Hermann15}. Additions they make for their model include a \emph{pointer-generator mechanism} \cite{Gulcehre16} that allows the model to copy words from the source document.
\citet{See17} propose a similar model, using a similar pointer-generator mechanism, that outperforms \citet{Nallapati16} on a slightly different version of the CNN/Daily Mail dataset (making the result not "strictly comparable"). They also incorporate what they call \emph{coverage} for avoiding repetitions in the output.


\section{Background}
\label{chap:theory}

In the following sections, various terms and concepts used throughout the paper are explained.

\subsection{Named Entity Recognition}
\label{sec:ner}

\emph{Information extraction} is a class of tasks that involve extracting structured information from documents. An example of such a task is \emph{named entity recognition}, which is the classification of parts of text into different categories, such as \emph{persons} or \emph{locations}, or no category. An example from the sentence "The mathematician Jeff Paris visited the city of Paris." is that "Jeff Paris" should be annotated as a person, and the last "Paris" as a location.

\subsection{Gated Recurrent Units}
The \emph{gated recurrent unit} (GRU) is a type of \emph{recurrent neural network} (RNN) that is designed to alleviate the \emph{vanishing/exploding gradient problem} \cite{Hochreiter91, Bengio94} which hinders the original RNN from capturing long term dependencies. GRU is similar to the popular \emph{long short-term memory} (LSTM) model but is simpler and less computationally intensive, while still achieving comparable results on many tasks \cite{Chung14, Kumar16}.
The entire GRU architecture can be described by the formulas
\begin{align*}
r_t &= \sigma(W^r[x_{t}, h_{t-1}] + b^r) \\
z_t &= \sigma(W^z[x_{t}, h_{t-1}] + b^z)
\\[0.3cm]
h'_{t} &= \tanh(W^h[x_{t}, r_t \odot h_{t-1}] + b^h)
\\
h_{t} &= z_t \odot h_{t-1} + (1-z_t) \odot h'_{t},
\end{align*}
The vectors $x_t$ is the input at time step $t$, and $h_t$ is the output, while $r_t$ and $z_t$ are scaling vectors, intended to regulate what information is let through. These can be described as \emph{gates}. They have elements in $[0,1]$. The vector $h'_{t}$ is rather intended to carry data. Its elements are in $[-1,1]$, generated from a network with a $\tanh$ activation function. We denote an entire GRU update step as
\begin{align*}
h_t = \GRU(h_{t-1}, x_t).
\end{align*}

\subsection{Word Embeddings}

Given a vocabulary $V$, we can encode each word uniquely using a \emph{one-hot} encoding. This gives a vector of length $|V|$ where every word in the vocabulary is mapped uniquely to some dimension, which a value of 1, while the other dimensions are 0. This vector can be transformed to an embedding for the word by multiplying it by an embedding matrix $W\sub{emb}$ of dimensionality $d\sub{emb} \times |V|$, where $d\sub{emb}$ is the word embedding dimensionality, commonly a hyperparameter in neural network models.
The intention is that the embeddings capture some characteristics of words, giving useful vector representations. For instance, two related words such as football and soccer may be expected to be close to each other in the vector space. Two methods for generating word embeddings are word2vec \cite{Mikolov13} and GloVe \cite{Pennington2014}.

\subsection{Attention}

For many problems, it has been found to be beneficial to use more of the RNN states than the final fixed-size hidden state. \emph{Attention} is a mechanism for allowing the model to access more information in the decoding process, by letting it identify relevant parts of the input and use the encoder hidden state at these locations. This technique has been used successfully for machine translation \cite{Bahdanau15} and image captioning \cite{Xu15}.

\section{Model}
\label{chap:model}

\begin{figure*}[ht]
	\centering
	\includegraphics[width=0.75\linewidth]{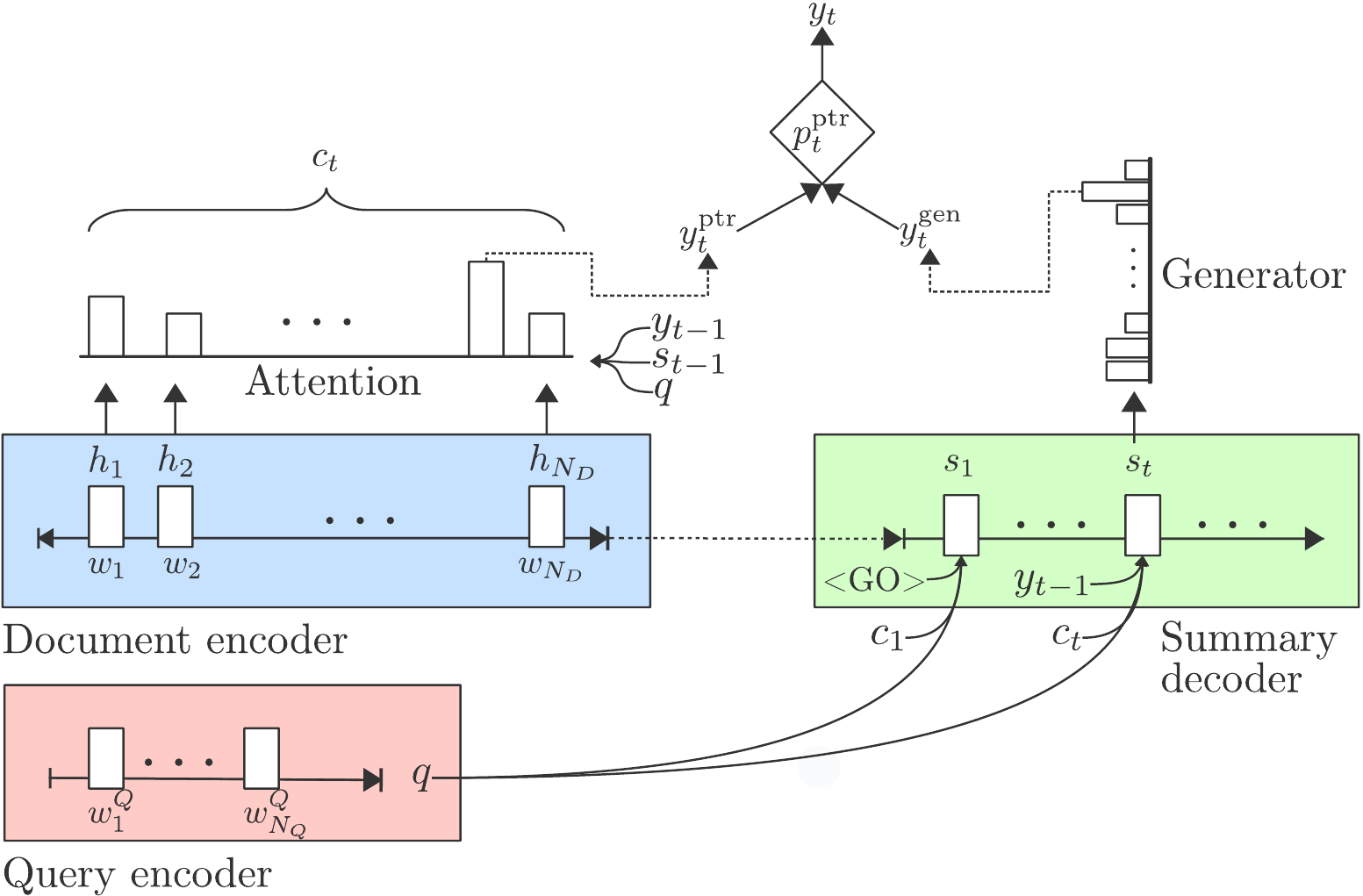}
	\caption{Overview of our model. It illustrates connections between parts of the model at a fixed decoder time step $t$. The bottom part, containing labeled boxes, correspond to the different RNNs. The top part is intended to visualize the two ways the output word $y_t$ can be selected, through the pointer and generator mechanism, to the left and right respectively.}
	\label{fig:model_overview}
\end{figure*}

We propose a sequence-to-sequence model with attention and a pointer mechanism, making it a \emph{pointer-generator} model. The input for the problem is a document and a query. These are sequences of words passed to a document encoder and a query encoder respectively. The encoders' outputs are then passed to the attentive decoder, which generates a summary. Both encoders, as well as the decoder, use RNNs with GRUs. Each occurrence of $\GRU$, with a subscript, in the formulas in the following sections has separate weights and biases. The entire model is depicted in Figure \ref{fig:model_overview}. The different components and variables in the figure will be explained in detail throughout the section.
\subsection{Document Encoder}
The document encoder processes an input document, generating a state for each input word. To get a representation of the context around a word, we use a bidirectional RNN \cite{Schuster97} encoder, so both the context before and after contribute to the representation.  This is used by \citet{Bahdanau15} amongst others, achieving good results on a similar task related to text comprehension.

The combined RNN hidden state at time step $i$, $h_i$, and the intermediate states, $\rarrowover{h}_i$ and $\larrowover{h}_i$, from the forward reader and backward reader respectively, are computed as
\begin{align*}
\rarrowover{h}_i &= \GRU_{\overrightarrow{\text{doc}}}(\rarrowover{h}_{i-1}, E(w_i)) \\
\larrowover{h}_i &= \GRU_{\overleftarrow{\text{doc}}}(\larrowover{h}_{i-1}  E(\larrowover{w}_i)) \\
h_i &= [\rarrowover{h}_i,\larrowover{h}_i],
\end{align*}
where $w_i\in V$, for the vocabulary $V$, is word $i$ in the input document; $\larrowover{w}_i$ is word $i$ in the reversed input; and $E(w_i)$ is the word embedding of $w_i$. The initial states $\rarrowover{h}_0$ and $\larrowover{h}_0$ are zero vectors. Due to the concatenation, the combined state $h_i$ has twice the dimensionality of the state of each unidirectional encoder. The document encoder state dimensionality is denoted $d\sub{doc}$ and the word embedding dimensionality $d\sub{emb}$.

\subsection{Query Encoder}
The query encoder is responsible for creating a fixed-size internal representation of the input query. Unlike the document encoder, the query encoder is a unidirectional RNN encoder since queries are relatively short compared to documents and we only use the final state to represent the whole query. The RNN state $h_i^Q$ at query word $i$, is updated according to $h^Q_i = \GRU\sub{que}(h^Q_{i-1}, E(w^Q_i)), q = h^Q_{N_Q}$,
where $w^Q$ is the input query and $N_Q$ is the length of the query. The initial state $h^Q_0$ is the zero vector. The query encoder state dimensionality is denoted $d\sub{que}$.
\subsection{Decoder}
The decoder is a unidirectional RNN for constructing a summary of the input document by depending on the final state of the input encoder, the query. It utilizes soft attention, in combination with a pointer mechanism, as well as a \emph{generator} part similar to \citet{Bahdanau15}. The query embedding $q$ is fed as input at each decoder time step. This is similar to the \emph{answering module} in a question answering model presented by \citet{Kumar16}, who use an RNN-encoded \emph{question} representation as input at each decoder time step. In our model, the RNN state is updated according to $s_t = \GRU\sub{dec}(s_{t-1}, [c_t, q, E(y_{t-1})])$,
where $s_0 = h_{N_D}$, the final document encoder state, $N_D$ being the number of input words; $y_0$ corresponds to a special \go token, used at the initial time step when no previous word has been predicted; $c_t$ is the \emph{context vector} at time step $t$ from the attention mechanism, defined subsequently; and $y_{t-1}\in V$ is the predicted output word at time step $t-1$. This is either from the generator mechanism, or the pointer mechanism, also defined subsequently. The word embeddings are the same as are used in the encoder.

The intention of the inclusion of $q$ to the input of $\GRU\sub{dec}$ is to give the decoder the ability to tune the structure of the output sequence to eventually output something concerning the query. For example, if the query is a location, the decoder can output words leading up to an appropriate inclusion of the location.

The generator outputs a word from a subset of the vocabulary $V\sub{gen} \subseteq V$ at each time step. The selection of the output words is done through a distribution of words in $V\sub{gen}$, computed through a softmax as $ p\sup{gen}_{tj}= \frac{\exp(z_{tj})}{\sum_{k} \exp(z_{tk})}$,
for $j \le |V\sub{gen}|$, an index uniquely mapped to a word $w \in V\sub{gen}$, and $z_{tj}$ as defined subsequently. Defining this as the probability $P\sup{gen}_t(w)$, we then select output word $y\sup{gen}_t$ with the highest probability by $ y\sup{gen}_t = \argmax{w \in V\sub{gen}} P\sup{gen}_t(w)$.
The softmax probability depends on $z_{tj}$, the output from two linear transformations on the decoder state and context vector, defined as $z_t = W\sub{gen}^{(2)} (W\sub{gen}^{(1)} [s_t, c_t] + b\sub{gen}^{(1)}) + b\sub{gen}^{(2)}$,
where $W\sub{gen}^{(1)} \in \R^{d\sub{gen} \times (d\sub{dec} + d\sub{doc})}$, $b\sub{gen}^{(1)}\in \R^{d\sub{gen}}$, $W\sub{gen}^{(2)}\in \R^{|V\sub{gen}| \times d\sub{gen}}$ and $b\sub{gen}^{(2)}\in \R^{|V\sub{gen}|}$ are trainable hyperparameters, in which $d\sub{gen}$ is the dimensionality of the hidden layer. The main function of this layer is to reduce the dimensionality of the input, for reducing computation time for the final layer with size $|V\sub{gen}|$.

The model has a soft attention mechanism, based on one used by \citet{Bahdanau15} for machine translation. The result of the attention mechanism is a context vector $c_t$ produced at each time step $t$, computed as
\begin{align*}
    c_t &= \sum_i \alpha_{ti} h_i \\
    \alpha_{ti} &= \frac{\exp(e_{ti})}{\sum_k \exp(e_{tk})} \\
    e_{ti} &= \text{score}(h_i, s_{t-1}, E(y_{t-1}), q)
\end{align*}
where $h_i$ is the document encoder hidden state at index $i$. The score function is defined as
    $\text{score}(h, s, x, q) = v\sub{att}^\intercal \tanh(W\sub{att} [h, s, x, q] + b\sub{att})$,
where $W\sub{att}\in \R^{d\sub{att} \times (d\sub{doc} + d\sub{dec} + d\sub{emb} + d\sub{que} )}$ is a weight matrix, $v\sub{att} \in \R^{d\sub{att}}$ is a vector, and $b\sub{att}$ is a bias vector, all of which are trained together with the rest of the network. The query $q$ is included for the model to focus attention around query words when appropriate.

\subsection{Pointer Mechanism}
\label{sec:pointer_mechanism}
A general issue is that with a generator mechanism limited to frequent words, infrequent words cannot be generated. Further, if the model needs to learn to output names, and there are many different ones and few occurrences of each in the training data, training a model to generate them correctly is problematic. A way to solve these issues is to allow the model to directly copy a word in the input document to the output summary, or \emph{point} to it. This may additionally be viewed as using the input text as a secondary output vocabulary, in addition to $V\sub{gen}$.

%
The pointer mechanism adds a \emph{switch}, $p\sup{ptr} \in (0,1)$, at each decoder time step $t$, to the model. It is computed as the output of a linear transformation fed through a sigmoid activation function, as
$p\sup{ptr}_t = \sigma(v\sub{ptr}^\intercal [s_t, E(y_{t-1}), c_t] + b\sub{ptr} )$,
where $v\sub{ptr} \in \R^{d\sub{dec} + d\sub{emb} + d\sub{doc}}$ and $b_{ptr}$ are vectors, all of which are trained together with the rest of the network.

If $p\sup{ptr}_t > 0.5$, a word is copied from the input, otherwise the generator output is used. What is copied from the input for the $t$th decoder word is determined by the attention distribution. Specifically, at time step $t$, we select the word at index
$i'_t = \argmax{i} \alpha_{ti}$
in the document, where the attention is highest, as
$y\sup{ptr}_t = w_{\left(i'_t\right)}$.
The final output word can then be defined as
\begin{align*}
y_t &= \begin{cases}
      y\sup{ptr}_t & \text{if $p\sup{ptr}_t > 0.5$} \\
      y\sup{gen}_t & \text{otherwise}
    \end{cases}
\end{align*}

\subsection{Training Loss}
The model is trained in when to use the pointer mechanism in a supervised manner. We define an additional training input $x\sup{ptr}_t$ that is either 1 if the pointer mechanism is set to be used for the $t$th word in the summary, or 0 otherwise. For training this, we define a loss function $L\sub{ptr} = \sum_{t=1}^{N_S} (x\sup{ptr}_t (-\log{p\sup{ptr}_t}) + (1-x\sup{ptr}_t) (-\log({1 - p\sup{ptr}_t})))$.

For training the generator mechanism, we define a loss over the generator softmax layer as
$L\sub{gen} = \sum_{t=1}^{N_S}(1-x\sup{ptr}_t) (-\log{P\sup{gen}_t(w^*)})$,
where $N_S$ is the length of the target summary, $w^* \in V\sub{gen}$ is the the $t$th word in the target summary. Multiplying by $(1-x\sup{ptr}_t) $ excludes any addition to the loss when the pointer mechanism is set to be used.

We introduce a form of supervised attention for when the pointer mechanism is set to be used for an output word by introducing a loss function
$L\sub{att} = \sum_{t=1}^{N_S}x\sup{ptr}_t (-\log{\alpha_{ti^*}})$,
where $i^*$ is the index in the input document to point to.

The final loss function is the sum of the different losses, normalized by the length, computed as
%
$L = \frac{1}{N_S}(L\sub{gen} + L\sub{att} + L\sub{ptr})$.

\subsection{Generating Summaries}
Summaries are considered complete when a special \eos token has been generated, or after a maximum output length is reached. Potential summaries are explored using beam search. 
However, for time steps where the pointer mechanism is used, the partial summaries are prioritized by probabilities as if the generator had been used instead, so $k$ partial summaries with different probabilities are created for the word chosen by the pointer mechanism. This is difficult to justify, but we hope that this should give a reasonable probability at time steps when the pointer mechanism is used, preventing summaries using the pointer mechanism more to be prioritized.

A slight deviation from what is presented in Section \ref{sec:pointer_mechanism} is that when the pointer mechanism is used and the attended word was not in $V$, we do not output \unk, which it is otherwise interpreted as in the model, but rather the actual input word before it being converted to an index in the vocabulary. This may be viewed as a post-processing step.
\section{Dataset}
\label{chap:dataset}
The dataset constructed for this paper is based \citet{Hermann15} and consist of document–query–answer triples from CNN and Daily Mail news articles. Included with each published news article, there are a number of human-written \emph{highlights}, which summarize different aspects of the article. Table \ref{tab:highlight_example} shows some example highlights for a single article. They construct a document–query–answer by considering a named entity in a highlight to be unknown, making the highlight into a Cloze-style question \cite{Taylor53}, whose answer is the entity made unknown. An example document and a Cloze-style question and its answer can be seen in Table \ref{tab:dataset_example}.
\begin{table}
\renewcommand{\arraystretch}{1.5}
\centering
\caption{Highlights of a CNN article titled "Airline quality report sorts out the duds from the dynamos in 2012".}
\begin{tabularx}{0.5\textwidth}{>{\footnotesize}X}
\toprule[0.3mm]
\raggedright\arraybackslash
\textbf{1.}
Hawaiian Airlines again lands at No. 1 in on-time performance\newline
\textbf{2.}
The Airline Quality Rankings Report looks at the 14 largest U.S. airlines\newline
\textbf{3.}
ExpressJet and American Airlines had the worst on-time performance\newline
\textbf{4.}
Virgin America had the best baggage handling; Southwest had lowest complaint rate \\
\bottomrule[0.3mm]
\end{tabularx}
\label{tab:highlight_example}
\end{table}
We propose using the CNN/Daily Mail dataset for query-based abstractive summarization by regarding each highlight as a summary of its document, and entities in the highlight as queries. For every occurrence of an entity in a highlight, we construct a document-query-summary triple for query-based summarization. Table \ref{tab:dataset_example} shows for a sample document a Cloze-style question compared and the corresponding query-summary pair constructed by us. If an entity is mentioned in multiple highlights, we consider there being multiple target references for the document-query pair. In contrast to \citet{Hermann15}, we do not translate entities into identifiers but use only minimal preprocessing in the form of tokenization and lowercasing. Further, we mix articles from DNN and Daily mail while \citet{Hermann15} keeps them separate. We decided to train our model on a mix of CNN and Daily Mail articles, with a proportion of them being reserved for validation and test sets. Which articles are included for the validation and test set is determined randomly with equal probability for every article. 

Some statistics of the resulting dataset can be seen in Table \ref{tab:dataset_statistics}.
\newcommand{\mypound}{\scalebox{0.8}{\raisebox{0.4ex}{\#}}}
\begin{table}
\renewcommand{\arraystretch}{1.1}
\centering
\caption{Statistics of the dataset.} 
\begin{tabularx}{.5\textwidth}{lrrr}
\toprule[0.3mm]
& \textbf{Training} & \textbf{Val.} & \textbf{Test}
\\
\hline
\mypound{}doc & 300,805 & 4,652 & 4,652 \\
\mypound{}doc-query pairs & 1,066,377 & 16,308 & 16,593 \\
\mypound{}doc-query-sum & 1,294,730 & 19,827 & 20,046 \\
avg \mypound{}words/doc & 773.02 & 778.78 & 775.70 \\
avg \mypound{}words/query & 1.52 & 1.53 & 1.52 \\
avg \mypound{}words/sum & 14.44 & 14.52 & 14.40
\\
\bottomrule[0.3mm]
\end{tabularx}
\label{tab:dataset_statistics}
\end{table}
The dataset can be reproduced using a script made available on GitHub\footnote{\url{https://github.com/helmertz/querysum-data}}.
\section{Experiments}
\label{chap:experiments}
Two experiments were conducted. The first to measure if the model uses the information in the query, \Sec{sec:query_dependence}, and the second compares the model to an extractive baseline, \Sec{sec:baseline}. A beam width of $k=5$ and a maximum output length of $32$ was used.
\subsection{Query Dependence}
\label{sec:query_dependence}
To determine whether incorporating a query benefits our model, we compare our proposed model to one where the query is corrupted. Instead of evaluating the generated summary for a document and a query with ID $n$ against the reference summaries for that query, we evaluate it against the reference summaries for query $n+1$, i.e. the query ID has been offset. For the query with the highest ID, the reference summaries for the first query are used. The idea is that if the score is lower than for the normal evaluation, then the model has made use of the additional information in the query. Table \ref{tab:query-offset-example} shows for an example document, 1, what the generated summaries are evaluated against during the query-dependence evaluation.
\begin{table}
\centering
\caption{Reference summaries used during normal evaluation compared to with offset queries.}
\begin{tabularx}{.48\textwidth}{@{}ccc@{}}
\toprule[0.3mm]
\textbf{Query ID} & \textbf{Normal} & \textbf{Offset queries}
\\
\hline
1.1 & A.1.1, B.1.1 & A.1.2 \\
1.2 & A.1.2 & A.1.3, B.1.3  \\
1.3 & A.1.3, B.1.3 & A.1.1, B.1.1  \\
\bottomrule[0.3mm]
\end{tabularx}
\label{tab:query-offset-example}
\end{table}
It is worth to mention that two reference summaries for different queries may be the same, as the same original highlight may be used as a reference summary for multiple queries. In these cases, the query will be appropriate for the summary and the model may have benefited from the query even in the query-offset evaluation.
\subsection{Extractive Baseline}
\label{sec:baseline}
As a baseline, we compare the results to a simple extractive summary, designed specifically for the dataset used in this thesis work. The baseline summary is constructed by selecting the first sentence in the document containing the query, without restricting the length of the document. If no such sentence is found, i.e. the document does not contain the query, the first sentence of the document is used instead. This does occur in the dataset, but not frequently.

We additionally observe that the average length of baseline sentences using the CNN/Daily Mail dataset is commonly greater than for the reference summaries. The average number of words is 30.56 for the baseline summaries, while it is 14.44 for the reference summaries. It may be possible to gain a higher ROUGE score if a fewer number of words around the query occurrence is selected, but it might not form a complete sentence.
\subsection{Evaluation Metric}
\label{sec:evaluation}
%
%
Our results are evaluated using four different metrics provided by \emph{ROUGE} (Recall-Oriented Understudy for Gisting Evaluation) \cite{Lin04}, the defacto standard evaluation method for automatic summarization. \emph{ROUGE-1}, \emph{ROUGE-2}, \emph{ROUGE-L}, and \emph{ROUGE-SU4}. ROUGE-1 and ROUGE-2 are the scores for 1-grams and 2-grams respectively. ROUGE-L and ROUGE-SU4 are more complex metrics, detailed by \citet{Lin04}. 

\subsection{Training Details}
\label{sec:training_details}
%
The vocabulary $V$ used for the input text contains the 150,000 most frequent words in the training set while the generator vocabulary $V\sub{gen}$ consist of the 20,000 most frequent words. The smaller vocabulary of the generator is due to the pointer mechanism.
%

Word embeddings for the vocabulary words are initialized with 100-dimensional GloVe embeddings\footnote{Downloadable as "glove.6B.zip" at: \url{https://nlp.stanford.edu/projects/glove/}}, trained on "Wikipedia 2014 + Gigaword 5". If the word does not have a GloVe embedding, we initialize the word embedding by sampling the per-dimension univariate normal distributions with means and standard deviations of the entire collection of GloVe embeddings. 

Both during training and test time, we limit the document length to the first 800 words, to reduce computation time.

The loss $L$ is minimized using the SGD-based Adam optimizer \cite{Kingma15}. We used mini-batches of 30 samples, with an averaged loss over all the samples in the batch. The mini-batches remained the same over epochs, but the order in which they were trained on was randomized between every epoch. 

Experiments have been run on a single Nvidia Tesla K80, with 12 GB of memory and took about 54 hours to train. The model is implemented using \emph{TensorFlow} \cite{Tensorflow2015}, and the complete source has been made available online\footnote{\url{https://github.com/helmertz/querysum}}. 

The hyperparameters used for the experiments is reported in Table \ref{tab:model_parameters}. No extensive hyperparameter tuning has been performed, but instead examined hyperparameters used for similar models, such as \citet{Nallapati16} and \citet{See17}.
\begin{table}[ht]
\caption{Hyperparameter configuration used.}
\begin{tabularx}{.48\textwidth}{llr}
\toprule[0.3mm]
\textbf{Hyperparameter} &  & \textbf{Value} \\
\hline
Word embedding size & $d\sub{emb}$ & 100 \\
Document encoder size & $d\sub{doc}$ & 512 \\
Query encoder size & $d\sub{que}$ & 256 \\
Decoder size & $d\sub{dec}$ & 512 \\
Attention hidden size & $d\sub{att}$ & 256 \\
Generator hidden size & $d\sub{gen}$ & 256 \\
\bottomrule[0.3mm]
\end{tabularx}
\label{tab:model_parameters}
\end{table}

\newpage

\section{Results}
\label{chap:results}
The results from our experiments are summarised in Table \ref{tab:rouge}.
\begin{table}
\centering
\caption{ROUGE scores of the evaluated models.}
\begin{tabular}{lrrrr}
\toprule[0.3mm]
{Model} & \multicolumn{1}{c}{{1}} & \multicolumn{1}{c}{{2}} & \multicolumn{1}{c}{{L}} & \multicolumn{1}{c}{{SU4}}
\\
\hline
\multicolumn{1}{m{1.7cm}}{{First query sentence}} & \textbf{33.81} & \textbf{18.19} & \textbf{29.22} & \textbf{17.49}
\\
Our model & 18.25 & 5.04 & 16.17 & 6.13
\\
Offset queries & 16.06 & 3.89 & 14.25 & 5.18
\\
\bottomrule[0.3mm]
\end{tabular}
\label{tab:rouge}
\end{table}
From the result of the query dependence evaluation ("offset queries"), described in Section \ref{sec:query_dependence}, we can see that the ROUGE scores goes down, with statistical significance according to the ROUGE-reported 95\% confidence intervals, when the queries are offset. This indicates that the model benefits from the information provided by queries.

Further, we observe that our model score lower than the baseline model which we denote the \emph{first query sentence} described in Section \ref{sec:baseline}. However, it should be noted that this baseline is expected to be strong given the nature of this dataset. 
%
%
\subsection{Further Analysis}
We observe that the attention at a time step appears to often be highly focused on only a few words in the document. An example of an output summary can be seen in Table \ref{tab:decode_example}, and Figure \ref{fig:attention_example1} shows the attention distribution over time for the same generated summary.
\begin{table}
\centering
\caption{Example document-query pair.}
\begin{tabularx}{.48\textwidth}{>{\footnotesize}X}
\toprule[0.3mm]
\textbf{Document}
( cnn ) -- the united states have named former germany captain jurgen klinsmann as their new national coach , just a day after sacking bob bradley . bradley , who took over as coach in january 2007 , was relieved of his duties on thursday , and u.s. soccer federation president sunil gulati confirmed in a statement on friday that his replacement has already been appointed . [...] 
\\
\textbf{Query}
united states
\\
\textbf{Reference}
jurgen klinsmann is named as coach of the united states national side
\\
\textbf{Output}
klinsmann appointed as the new coach of united states
\\
\bottomrule[0.3mm]
\end{tabularx}
\label{tab:decode_example}
\end{table}
%
%
Another observation we make is that the attention often is focused at the beginning of the documents. However, there are certainly instances when entities are selected from far back in documents. This bias may partly be due to our decision to point out the first occurrences of entities. Although, it has been noted by \citet{Goldstein99} that the beginning of news articles often summarizes the article quite well.

From examining some of the output summaries from our model, we see that they often strongly match the topic of the input documents, but they rarely succeed in generating summaries rephrasing something actually stated in the article. Table \ref{tab:falseexample} shows an example output that is fairly grammatically correct, but not truthful with respect to the article.
%
\begin{table}
\centering
\caption{Example document-query pair.}
\begin{tabularx}{.48\textwidth}{>{\footnotesize}X}
\toprule[0.3mm]
\textbf{Document}
president barack obama sided with open-internet activists on monday , urging the federal communications commission to draft new rules that would reclassify the broadband net to regulate it more like a public utility . the end result would tie the hands of internet service providers that want to cut special deals with services like netflix , youtube , hulu and amazon to push their streaming content along a ' fast lane ' that ordinary americans ca n't access . [...]
\\
\textbf{Query}
netflix
\\
\textbf{Reference}
obama 's vision would bar providers like verizon and comcast from cutting deals with hulu , netflix and amazon so their streaming content could be delivered along online ' fast lanes '\\
\textbf{Output}
obama 's chief executive of netflix has refused to allow users to access the service
\\
\bottomrule[0.3mm]
\end{tabularx}
\label{tab:falseexample}
\end{table}

We observe that the model manages to learn some of the dataset samples which are not actual summaries, described in Section \ref{chap:dataset}, such as notices repeated over several articles. The generated summary shown in Table \ref{tab:decode_artifact_example} is an example of this. Interestingly, the model manages to literally repeat the reference summary, up to the maximum output length limit.
%
\begin{table}
\centering
\caption{Example document-query pair.}
\begin{tabularx}{.48\textwidth}{>{\footnotesize}X}
\toprule[0.3mm]
\textbf{Document}
february 13 , 2015 a breakthrough in belarus , a verdict in italy , and an expected veto in the u.s. all headline cnn student news this friday . [...] 
\\
\textbf{Query}
cnn student news roll call
\\
\textbf{Reference}
at the bottom of the page , comment for a chance to be mentioned on cnn student news . you must be a teacher or a student age 13 or older to request a mention on the cnn student news roll call .
\\
\textbf{Output}
at the bottom of the page , comment for a chance to be mentioned on cnn student news . you must be a teacher or a student age 13 or older to
\\
\bottomrule[0.3mm]
\end{tabularx}
\label{tab:decode_artifact_example}
\end{table}
We can frequently see repetitions of the same phrases; an extreme example can be seen in Figure \ref{fig:attention_example2}. The model appears to get stuck trying to begin a summary. Additionally, we observe that the repetition can be observed in the attention distribution as well. The same problem has been seen by \citet{Nallapati16}, who make an addition, \emph{temporal attention} \cite{Sankaran16}, to their model for alleviating the issue of repetitions. \citet{See17} propose using \emph{coverage} to solve the same issue. 
%

Before running experiments, we suspected that it may be difficult for the pointer mechanism to sequentially point out words that make up longer entities. However, we see that this is done successfully quite often. For an example summary, the certainty of selecting a sequence of entity words can be seen in Figure \ref{fig:longpointexample}.
%

Compared to the reference summaries, the output is generally shorter. The average number of words in output summaries is 11.27, while the dataset average is 14.44. As is noted by \citet{Wu16}, beam search commonly favors shorter summaries. They propose an addition of \emph{length normalization}, for reducing this tendency. Implementing such a measure may improve the results of our model as well.

In comparison to \citet{Nallapati16} and \citet{See17}, our ROUGE scores are low. They use a different version of the dataset where all highlights are combined to form a single, often multi-sentence, summary. With similar models, they get ROUGE-1 results of around 35 on the general summarization task. 
However, while they always train the model to output the same summary for the same document, we often have completely different target summaries for different queries, where the queries make up a much smaller part of the input. 
\flushbottom

\section{Conclusion}
\label{chap:conclusion}

We have designed a model for query-based abstractive summarization and evaluated it on an adapted QA dataset, redesigned for query-based summarization. While the overall performance of the model is not enough to outperform our extractive baseline, we have shown that it can incorporate a query and utilize the information to create more focused summaries.

\bibliography{references}

\begin{thebibliography}{}
\expandafter\ifx\csname natexlab\endcsname\relax\def\natexlab#1{#1}\fi

\bibitem[{Abadi et~al.(2015)Abadi, Agarwal, Barham, Brevdo, Chen, Citro,
  Corrado, Davis, Dean, Devin, Ghemawat, Goodfellow, Harp, Irving, Isard, Jia,
  Jozefowicz, Kaiser, Kudlur, Levenberg, Man{\'e}, Monga, Moore, Murray, Olah,
  Schuster, Shlens, Steiner, Sutskever, Talwar, Tucker, Vanhoucke, Vasudevan,
  Vi{\'e}gas, Vinyals, Warden, Wattenberg, Wicke, Yu, and
  Zheng}]{Tensorflow2015}
Mart{\'i}n Abadi, Ashish Agarwal, Paul Barham, Eugene Brevdo, Zhifeng Chen,
  Craig Citro, Greg Corrado, Andy Davis, Jeffrey Dean, Matthieu Devin, Sanjay
  Ghemawat, Ian Goodfellow, Andrew Harp, Geoffrey Irving, Michael Isard,
  Yangqing Jia, Rafal Jozefowicz, Lukasz Kaiser, Manjunath Kudlur, Josh
  Levenberg, Dan Man{\'e}, Rajat Monga, Sherry Moore, Derek Murray, Chris Olah,
  Mike Schuster, Jonathon Shlens, Benoit Steiner, Ilya Sutskever, Kunal Talwar,
  Paul Tucker, Vincent Vanhoucke, Vijay Vasudevan, Fernanda Vi{\'e}gas, Oriol
  Vinyals, Pete Warden, Martin Wattenberg, Martin Wicke, Yuan Yu, and Xiaoqiang
  Zheng. 2015.
\newblock {TensorFlow}: Large-scale machine learning on heterogeneous systems.
\newblock Software available from
  \href{https://www.tensorflow.org/}{\texttt{tensorflow.org}}.

\bibitem[{Bahdanau et~al.(2015)Bahdanau, Cho, and Bengio}]{Bahdanau15}
Dzmitry Bahdanau, Kyunghyun Cho, and Yoshua Bengio. 2015.
\newblock Neural machine translation by jointly learning to align and
  translate.
\newblock {\em International Conference on Learning Representations (ICLR
  2015)\/} arXiv:1409.0473.

\bibitem[{Bengio et~al.(1994)Bengio, Simard, and Frasconi}]{Bengio94}
Yoshua Bengio, Patrice Simard, and Paolo Frasconi. 1994.
\newblock Learning long-term dependencies with gradient descent is difficult.
\newblock {\em IEEE transactions on neural networks\/} 5(2):157--166.

\bibitem[{Chung et~al.(2014)Chung, G{\"{u}}l{\c{c}}ehre, Cho, and
  Bengio}]{Chung14}
Junyoung Chung, {\c{C}}aglar G{\"{u}}l{\c{c}}ehre, KyungHyun Cho, and Yoshua
  Bengio. 2014.
\newblock Empirical evaluation of gated recurrent neural networks on sequence
  modeling.
\newblock {\em ArXiv e-prints\/} arXiv:1412.3555.

\bibitem[{Goldstein et~al.(1999)Goldstein, Kantrowitz, Mittal, and
  Carbonell}]{Goldstein99}
Jade Goldstein, Mark Kantrowitz, Vibhu Mittal, and Jaime Carbonell. 1999.
\newblock Summarizing text documents: Sentence selection and evaluation
  metrics.
\newblock In {\em Proceedings of the 22Nd Annual International ACM SIGIR
  Conference on Research and Development in Information Retrieval\/}. ACM, New
  York, NY, USA, SIGIR '99, pages 121--128.

\bibitem[{G{\"{u}}l{\c{c}}ehre et~al.(2016)G{\"{u}}l{\c{c}}ehre, Ahn,
  Nallapati, Zhou, and Bengio}]{Gulcehre16}
{\c{C}}aglar G{\"{u}}l{\c{c}}ehre, Sungjin Ahn, Ramesh Nallapati, Bowen Zhou,
  and Yoshua Bengio. 2016.
\newblock Pointing the unknown words.
\newblock In {\em Proceedings of the 54th Annual Meeting of the Association for
  Computational Linguistics (Volume 1: Long Papers)\/}. Association for
  Computational Linguistics, Berlin, Germany, pages 140--149.

\bibitem[{Hermann et~al.(2015)Hermann, Kocisk{\'{y}}, Grefenstette, Espeholt,
  Kay, Suleyman, and Blunsom}]{Hermann15}
Karl~Moritz Hermann, Tom{\'{a}}s Kocisk{\'{y}}, Edward Grefenstette, Lasse
  Espeholt, Will Kay, Mustafa Suleyman, and Phil Blunsom. 2015.
\newblock Teaching machines to read and comprehend.
\newblock In {\em Advances in Neural Information Processing Systems\/}. pages
  1693--1701.

\bibitem[{Hochreiter(1991)}]{Hochreiter91}
Sepp Hochreiter. 1991.
\newblock {\em Untersuchungen zu dynamischen neuronalen Netzen\/}.
\newblock Ph.D. thesis, diploma thesis, institut f{\"u}r informatik, lehrstuhl
  prof. brauer, technische universit{\"a}t m{\"u}nchen.

\bibitem[{Karpathy and Fei-Fei(2015)}]{Karpathy15}
Andrej Karpathy and Li~Fei-Fei. 2015.
\newblock Deep visual-semantic alignments for generating image descriptions.
\newblock In {\em The IEEE Conference on Computer Vision and Pattern
  Recognition (CVPR)\/}.

\bibitem[{Kingma and Ba(2015)}]{Kingma15}
Diederik~P. Kingma and Jimmy Ba. 2015.
\newblock Adam: {A} method for stochastic optimization.
\newblock {\em International Conference on Learning Representations (ICLR
  2015)\/} arXiv:1412.6980.

\bibitem[{Kumar et~al.(2016)Kumar, Irsoy, Ondruska, Iyyer, Bradbury, Gulrajani,
  Zhong, Paulus, and Socher}]{Kumar16}
Ankit Kumar, Ozan Irsoy, Peter Ondruska, Mohit Iyyer, James Bradbury, Ishaan
  Gulrajani, Victor Zhong, Romain Paulus, and Richard Socher. 2016.
\newblock Ask me anything: Dynamic memory networks for natural language
  processing.
\newblock In Maria~Florina Balcan and Kilian~Q. Weinberger, editors, {\em
  Proceedings of The 33rd International Conference on Machine Learning\/}.
  PMLR, New York, New York, USA, volume~48 of {\em Proceedings of Machine
  Learning Research\/}, pages 1378--1387.

\bibitem[{Lin(2004)}]{Lin04}
Chin-Yew Lin. 2004.
\newblock Rouge: A package for automatic evaluation of summaries.
\newblock In Stan~Szpakowicz Marie-Francine~Moens, editor, {\em Text
  Summarization Branches Out: Proceedings of the ACL-04 Workshop\/}.
  Association for Computational Linguistics, Barcelona, Spain, pages 74--81.

\bibitem[{Mikolov et~al.(2013)Mikolov, Sutskever, Chen, Corrado, and
  Dean}]{Mikolov13}
Tomas Mikolov, Ilya Sutskever, Kai Chen, Greg Corrado, and Jeffrey Dean. 2013.
\newblock Distributed representations of words and phrases and their
  compositionality.
\newblock In {\em Proceedings of the 26th International Conference on Neural
  Information Processing Systems\/}. Curran Associates Inc., USA, NIPS'13,
  pages 3111--3119.

\bibitem[{Mogren et~al.(2015)Mogren, K{\aa}geb{\"a}ck, and
  Dubhashi}]{mogren2015extractive}
Olof Mogren, Mikael K{\aa}geb{\"a}ck, and Devdatt~P Dubhashi. 2015.
\newblock Extractive summarization by aggregating multiple similarities.
\newblock In {\em RANLP\/}. pages 451--457.

\bibitem[{Nallapati et~al.(2016)Nallapati, Zhou, {Nogueira dos Santos},
  G{\"{u}}l{\c{c}}ehre, and Xiang}]{Nallapati16}
Ramesh Nallapati, Bowen Zhou, C{\'i}cero {Nogueira dos Santos}, {\c{C}}aglar
  G{\"{u}}l{\c{c}}ehre, and Bing Xiang. 2016.
\newblock Abstractive text summarization using sequence-to-sequence rnns and
  beyond.
\newblock In {\em Proceedings of the 20th {SIGNLL} Conference on Computational
  Natural Language Learning, CoNLL 2016, Berlin, Germany, August 11-12,
  2016\/}. pages 280--290.

\bibitem[{Nema et~al.(2017)Nema, Khapra, Laha, and Ravindran}]{Nema17}
Preksha Nema, Mitesh Khapra, Anirban Laha, and Balaraman Ravindran. 2017.
\newblock {Diversity driven Attention Model for Query-based Abstractive
  Summarization}.
\newblock {\em ArXiv e-prints\/} arXiv:1704.08300.

\bibitem[{Nenkova and McKeown(2012)}]{Nenkova12}
Ani Nenkova and Kathleen McKeown. 2012.
\newblock {\em A Survey of Text Summarization Techniques\/}, Springer US,
  Boston, MA, pages 43--76.

\bibitem[{Otterbacher et~al.(2009)Otterbacher, Erkan, and
  Radev}]{Otterbacher09}
Jahna Otterbacher, Gunes Erkan, and Dragomir~R. Radev. 2009.
\newblock Biased lexrank: Passage retrieval using random walks with
  question-based priors.
\newblock {\em Information Processing and Management\/} 45(1):42--54.

\bibitem[{Pennington et~al.(2014)Pennington, Socher, and
  Manning}]{Pennington2014}
Jeffrey Pennington, Richard Socher, and Christopher~D. Manning. 2014.
\newblock Glove: Global vectors for word representation.
\newblock In {\em Empirical Methods in Natural Language Processing (EMNLP)\/}.
  pages 1532--1543.

\bibitem[{Rush et~al.(2015)Rush, Chopra, and Weston}]{Rush15}
Alexander~M. Rush, Sumit Chopra, and Jason Weston. 2015.
\newblock A neural attention model for abstractive sentence summarization.
\newblock In {\em Proceedings of the 2015 Conference on Empirical Methods in
  Natural Language Processing\/}. Association for Computational Linguistics,
  Lisbon, Portugal, pages 379--389.

\bibitem[{Sankaran et~al.(2016)Sankaran, Mi, Al{-}Onaizan, and
  Ittycheriah}]{Sankaran16}
Baskaran Sankaran, Haitao Mi, Yaser Al{-}Onaizan, and Abe Ittycheriah. 2016.
\newblock Temporal attention model for neural machine translation.
\newblock {\em ArXiv e-prints\/} arXiv:1608.02927.

\bibitem[{Schuster and Paliwal(1997)}]{Schuster97}
Mike Schuster and Kuldip~K Paliwal. 1997.
\newblock Bidirectional recurrent neural networks.
\newblock {\em IEEE Transactions on Signal Processing\/} 45(11):2673--2681.

\bibitem[{See et~al.(2017)See, Liu, and Manning}]{See17}
Abigail See, Peter~J. Liu, and Christopher~D. Manning. 2017.
\newblock {Get To The Point: Summarization with Pointer-Generator Networks}.
\newblock {\em ArXiv e-prints\/} arXiv:1704.04368.

\bibitem[{Sutskever et~al.(2014)Sutskever, Vinyals, and Le}]{Sutskever14}
Ilya Sutskever, Oriol Vinyals, and Quoc~V. Le. 2014.
\newblock Sequence to sequence learning with neural networks.
\newblock In {\em Proceedings of the 27th International Conference on Neural
  Information Processing Systems\/}. MIT Press, Cambridge, MA, USA, NIPS'14,
  pages 3104--3112.

\bibitem[{Tan et~al.(2015)Tan, Xiang, and Zhou}]{Tan15}
Ming Tan, Bing Xiang, and Bowen Zhou. 2015.
\newblock Lstm-based deep learning models for non-factoid answer selection.
\newblock {\em ArXiv e-prints\/} arXiv:1511.04108.

\bibitem[{Taylor(1953)}]{Taylor53}
Wilson~L Taylor. 1953.
\newblock `cloze procedure': a new tool for measuring readability.
\newblock {\em Journalism Bulletin\/} 30(4):415--433.

\bibitem[{Wang et~al.(2013)Wang, Raghavan, Castelli, Florian, and
  Cardie}]{Wang13}
Lu~Wang, Hema Raghavan, Vittorio Castelli, Radu Florian, and Claire Cardie.
  2013.
\newblock A sentence compression based framework to query-focused
  multi-document summarization.
\newblock In {\em ACL 2013\/}.

\bibitem[{Wu et~al.(2016)Wu, Schuster, Chen, Le, Norouzi, Macherey, Krikun,
  Cao, Gao, Macherey, Klingner, Shah, Johnson, Liu, Kaiser, Gouws, Kato, Kudo,
  Kazawa, Stevens, Kurian, Patil, Wang, Young, Smith, Riesa, Rudnick, Vinyals,
  Corrado, Hughes, and Dean}]{Wu16}
Yonghui Wu, Mike Schuster, Zhifeng Chen, Quoc~V. Le, Mohammad Norouzi, Wolfgang
  Macherey, Maxim Krikun, Yuan Cao, Qin Gao, Klaus Macherey, Jeff Klingner,
  Apurva Shah, Melvin Johnson, Xiaobing Liu, Lukasz Kaiser, Stephan Gouws,
  Yoshikiyo Kato, Taku Kudo, Hideto Kazawa, Keith Stevens, George Kurian,
  Nishant Patil, Wei Wang, Cliff Young, Jason Smith, Jason Riesa, Alex Rudnick,
  Oriol Vinyals, Greg Corrado, Macduff Hughes, and Jeffrey Dean. 2016.
\newblock {Google's Neural Machine Translation System: Bridging the Gap between
  Human and Machine Translation}.
\newblock {\em ArXiv e-prints\/} arXiv:1609.08144.

\bibitem[{Xu et~al.(2015)Xu, Ba, Kiros, Cho, Courville, Salakhudinov, Zemel,
  and Bengio}]{Xu15}
Kelvin Xu, Jimmy Ba, Ryan Kiros, Kyunghyun Cho, Aaron Courville, Ruslan
  Salakhudinov, Rich Zemel, and Yoshua Bengio. 2015.
\newblock Show, attend and tell: Neural image caption generation with visual
  attention.
\newblock In {\em International Conference on Machine Learning\/}. pages
  2048--2057.

\end{thebibliography}
\bibliographystyle{acl_natbib}

\appendix
\renewcommand\thefigure{\thesection.\arabic{figure}} 
\setcounter{figure}{0} 
\renewcommand\thetable{\thesection.\arabic{table}} 
\setcounter{table}{0}
\section{Supplemental Material}

\subsection{Dataset}
An example of a record in the dataset is shown in Table \ref{tab:dataset_example}.

We organize the dataset triples hierarchically, first by document, then query, then reference. The documents and queries are numbered numerically starting with \emph{1}, while the references are numbered alphabetically starting with \emph{A}. Document 1 may have queries 1.1 and 1.2, and reference summaries A.1.1, B.1.1 and A.1.2\footnote{Selected for matching the format expected by pyrouge}. The order is shuffled amongst document, query and reference IDs.

\onecolumn
\begin{minipage}{\textwidth}
\subsection{Attention Visualisations}
\label{ap:Attention}
\begin{figure}[H]
    \centering
	\includegraphics[width=0.95\linewidth]{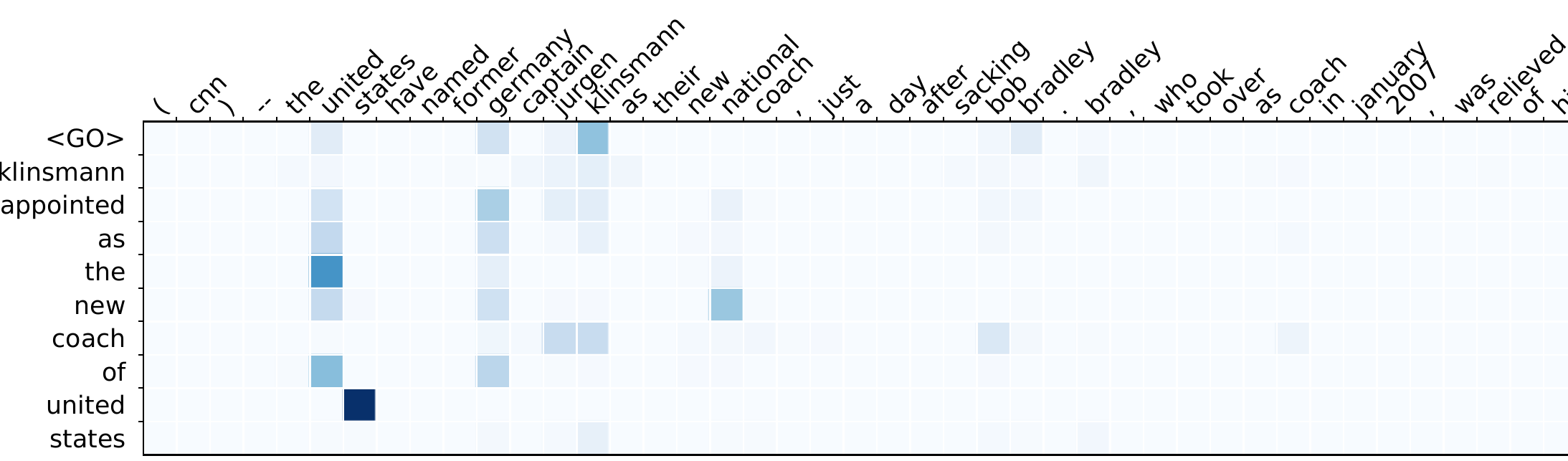} 
	\caption{Visualization of the attention distribution as the summary in Table \ref{tab:decode_example} is generated. The words of the document are shown on the horizontal axis, from left to right. Only a limited number of document words are shown. The vertical axis shows the output words, from top to bottom, after the \go token. The darker a cell is, the higher the attention on that position.}
	\label{fig:attention_example1}
	\includegraphics[width=0.95\linewidth]{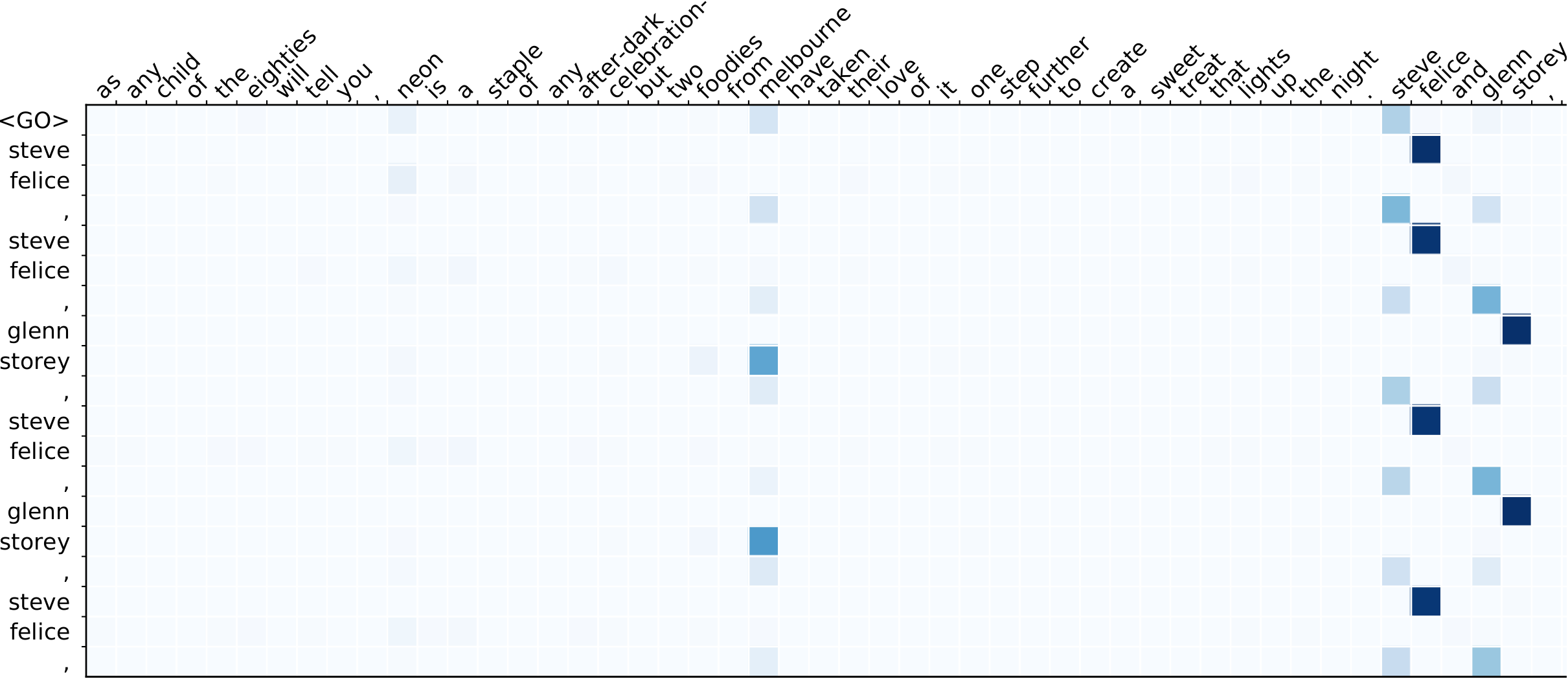} 
	\caption{Visualization of the attention distribution, $\alpha_{ti}$, as an output summary for a document-query pair is generated. The query is "australia". The format is the same as in Figure \ref{fig:attention_example1}.}
	\label{fig:attention_example2}
	\includegraphics[width=0.95\linewidth]{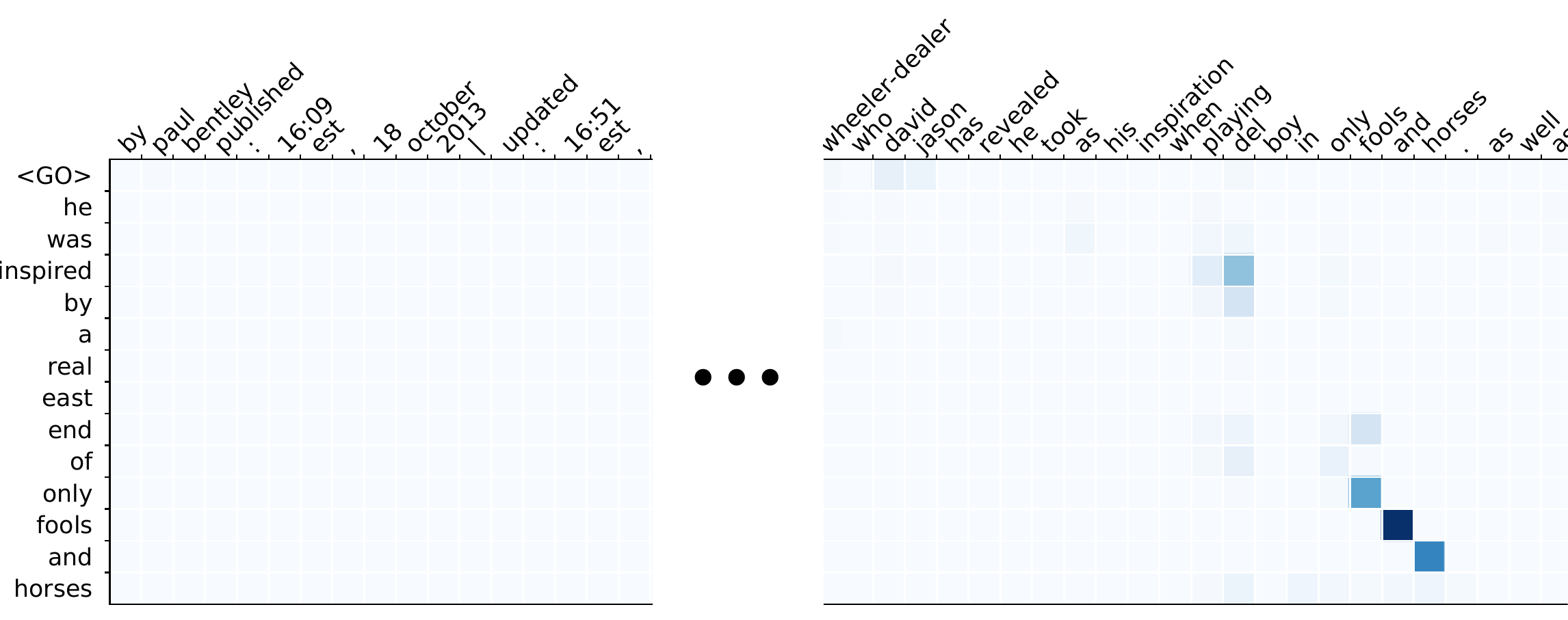} 
	\caption{Visualization of the attention distribution, $\alpha_{ti}$, as an output summary for a document-query pair in the test set is generated. The query is "only fools and horses". The format is the same as in Figure \ref{fig:attention_example1}. The ellipsis signifies that parts of the attention distribution has been skipped.}
	\label{fig:longpointexample}
\end{figure}
\end{minipage}

\twocolumn
\begin{table}
\renewcommand{\arraystretch}{1.5}
\centering
\caption{Example of dataset samples generated from a document-query pair using our method compared to \citet{Hermann15}. In the Cloze-style questions, the entity corresponding to the answer has been replaced by X.}
\begin{tabularx}{.5\textwidth}{>{\footnotesize}X}
\toprule[0.3mm]
\textbf{Document}
\newline
( cnn ) former vice president walter mondale was released from the mayo clinic on saturday after being admitted with influenza , hospital spokeswoman kelley luckstein said . `` he 's doing well . we treated him for flu and cold symptoms and he was released today , '' she said . mondale , 87 , was diagnosed after he went to the hospital for a routine checkup following a fever , former president jimmy carter said friday . `` he is in the bed right this moment , but looking forward to come back home , '' carter said during a speech at a nobel peace prize forum in minneapolis . `` he said tell everybody he is doing well . '' mondale underwent treatment at the mayo clinic in rochester , minnesota . the 42nd vice president served under carter between 1977 and 1981 , and later ran for president , but lost to ronald reagan . but not before he made history by naming a woman , u.s. rep. geraldine a. ferraro of new york , as his running mate . before that , the former lawyer was a u.s. senator from minnesota . his wife , joan mondale , died last year .
\\
\hline
\textbf{Highlight}
\newline
walter mondale was released from the mayo clinic on saturday , hospital spokeswoman said
\\
\hline
\textbf{Cloze-style question}
\newline
walter mondale was released from the X on saturday , hospital spokeswoman said
\\
\textbf{Cloze-style answer}
\newline
mayo clinic
\\
\hline
\textbf{Our query}
\newline
mayo clinic
\\
\textbf{Our target summary}
\newline
walter mondale was released from the mayo clinic on saturday , hospital spokeswoman said
\\
\bottomrule[0.3mm]
\end{tabularx}
\label{tab:dataset_example}
\end{table}

\end{document}